# Bayesian Semantic Instance Segmentation in Open Set World

Trung Pham, Vijay Kumar B G, Thanh-Toan Do,
Gustavo Carneiro, and Ian Reid

School of Computer Science, The University of Adelaide,
{trung.pham,vijay.kumar,thanh-toan.do,
gustavo.carneiro,ian.reid}@adelaide.edu.au

**Abstract.** This paper addresses the semantic instance segmentation task in the open-set conditions, where input images can contain known and unknown object classes. The training process of existing semantic instance segmentation methods requires annotation masks for all object instances, which is expensive to acquire or even infeasible in some realistic scenarios, where the number of categories may increase boundlessly. In this paper, we present a novel open-set semantic instance segmentation approach capable of segmenting all known and unknown object classes in images, based on the output of an object detector trained on known object classes. We formulate the problem using a Bayesian framework, where the posterior distribution is approximated with a simulated annealing optimization equipped with an efficient image partition sampler. We show empirically that our method is competitive with state-of-the-art supervised methods on known classes, but also performs well on unknown classes when compared with unsupervised methods.

## 1  Introduction

In recent years, scene understanding driven by multi-class semantic segmentation [10,13,16], object detection [19] or instance segmentation [7] has progressed significantly thanks to the power of deep learning. However, a major limitation of these deep learning based approaches is that they only work for a set of known object classes that are used during supervised training. In contrast, autonomous systems often operate under *open-set* conditions [23] in many application domains, i.e. they will inevitably encounter object classes that were not part of the training dataset. For instance, state-of-the-art methods such as Mask-RCNN [7] and YOLO9000 [19] fail to detect such unknown objects. This behavior is detrimental to the performance of autonomous systems that would ideally need to understand scenes holistically, i.e., reasoning about all objects that appear in the scene and their complex relations.

Semantic instance segmentation based scene understanding has recently attracted the interest of the field [3, 25]. The ultimate goal is to decompose the input image into individual objects (e.g., car, human, chair) and stuffs (e.g., road, floor) along with their semantic labels. Compared with semantic segmentation and object detection, the accuracy and robustness of semantic instance segmentation lags significantly. Recent efforts (e.g., [7]) follow a *detect-and-segment* approach — first detect objects in an image, then generate a segmentation mask for each instance. Such an approach might label a pixel with multiple object instances, and completely fails to segment unknown objects, and even known, but miss-detected objects. More importantly, current instance segmentation methods require annotation masks for all object instances during training, which is too expensive to acquire for new classes. A much cheaper alternative consists of the bounding box annotation of new classes (a mere two mouse clicks, compared to the multiple clicks required for annotating segmentation masks).

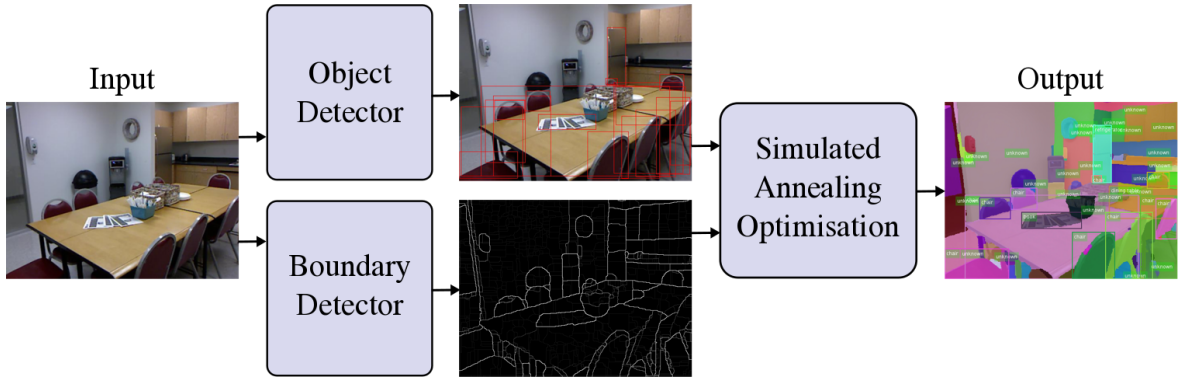

**Fig. 1:** Overview of semantic instance segmentation in a open-set environment. Our method segments all image regions irrespective of whether they have been detected or undetected, or are from a known or unknown class

In this paper, we propose a novel Bayesian semantic instance segmentation approach that is capable of segmenting all object instances irrespective of whether they have been detected or undetected and are from a known or an unknown training class. Such a capability is vitally useful for many vision-based robotic systems. Our proposed approach generates a global pixelwise image segmentation conditioned on a set of detections of known object classes (in terms of either bounding boxes or masks) instead of generating a segmentation mask for each detection (e.g., [7]). The segmentation produced by our approach not only keeps the benefits of the ability to segment known objects, but also retains the generality of an approach that can handle unknown objects via perceptual grouping. The outcome of our algorithm is a set of regions which are perceptually grouped and are each associated either to a known (object) detection or an unknown object class. To best of our knowledge, such a segmentation output has never been achieved before.

We formulate the instance segmentation problem using a Bayesian framework, where the likelihood is measured using image boundaries, a geometric bounding box model for pixel locations and optionally a mask model. These models compete with each other to explain different image regions. Intuitively, the boundary model explains unknown regions while bounding box and mask models describe regions where known objects are detected. The prior model simply penalizes the number of regions and enforces object compactness.

Nonetheless, finding the segmentation that maximizes the posterior distribution over a very large image partition space is non-trivial. Gibbs sampling [9] could be employed but it might take too long to converge. One of the main contributions of this work is an efficient image partition sampler that quickly generates high-quality segmentation proposals. Our image partition sampler is based on a boundary-driven region hierarchy, where regions of the hierarchy are likely representations of object instances. The boundary is estimated using a deep neural network [12]. To sample a new image partition, we simply select one region of the hierarchy, and "paste" it to the current segmentation. This operation will automatically realize either the split, merge or split-and-merge move between different segmentations depending on the selected region. Finally, the image partitioner is equipped with a Simulated Annealing optimization [28] to approximate the optimal segmentation.

We evaluate the effectiveness of our open-set instance segmentation approach on several datasets including indoor NYU [24] and general COCO [11]. Experimental results confirm that our segmentation method, with only bounding box supervision, is competitive with the state-of-the-art supervised instance segmentation methods (e.g., [7,8]) when tested on known object classes, while it is able to segment miss-detected and unknown objects. Our segmentation approach also outperforms other unsupervised segmentation methods when tested on unknown classes. Figure 1 demonstrates an overview and an example outcome of our segmentation method.

## 2   Related Work

**Supervised instance segmentation:** State-of-the-art supervised instance segmentation methods (e.g., [4,7,29]) follow a detect-and-segment approach — first detect objects in an image, then generate a segmentation mask for each instance. For example, the Mask-RCNN method [7] extends the Faster-RCNN [21] object detection network by adding another semantic segmentation branch for predicting a segmentation mask for each detected instance. Earlier methods [17,18] are based on segment proposals. For instance, DeepMask [17] and SharpMask [18] learn to generate segment proposals which are then classified into semantic categories using Fast-RCNN. In contrast, the FCIS method [29] jointly predicts, for each location in the image, an object class, a bounding box and a segmentation mask. The methods in [20, 22] employ Recurrent Neural Networks (RNN) to sequentially predict an object binary mask at each step.

Another group of supervised instance segmentation methods is based on clustering. In [5], the idea is first computing the likelihood that two pixels belong to the same object (using a deep neural network), then use these likelihoods to segment the image into object instances. Instead of predicting similarities between pixels, the method in [2] predicts a energy value for each pixel, the energy sur-face is then used to partition the image into object instances using the watershed transform algorithm.

The common drawback of existing instance segmentation methods is that they require a strong supervisory signal, consisting of the annotation masks of the known objects that are used during training. In contrast, our Bayesian instance segmentation approach does not necessarily require such object annotation masks, while being capable of segmenting all object instances irrespective of whether they have been detected or not and are from a known or unknown class.

**Unsupervised segmentation:** In contrast to learning based segmentation, unsupervised segmentation methods [6, 15, 26] are able to discover unknown objects without the strong supervisory training signal mentioned above. These methods, however, often make strong assumptions about visual objects (e.g., they tend to have similar color, texture and share strong edges) and consequently rely on low-level image cues such as color, depth, texture and edges for segmentation. As a result, their results tend to be relatively inaccurate. In contrast, our segmentation approach combines the best of both worlds using a unified formulation. In particular, our method exploits the prior object locations (for example given by an object detector) to improve the overall image segmentation. At the same time, our method does not require expensive segmentation masks of all object instances for training.

## 3   Open-set Semantic Instance Segmentation

Let $\mathbf{I}: \Omega \to \mathbb{R}$ be an input image defined on a discrete pixel grid $\Omega = \{v_1, v_2, \dots\}$, i.e., $\mathbf{I}_v$ is the color or intensity at pixel $v$. The goal of semantic instance segmentation is to decompose the image $\mathbf{I}_\Omega$ into individual object instance regions (e.g., chair, monitor) and stuff regions (e.g., floor, ceiling) along with their semantic labels. In particular, one seeks a partition of the image into $k$ *non-overlap* regions

$$\cup_{i=1}^{k} R_i = \Omega, \quad R_i \cap R_j = \emptyset, \; \forall i \neq j, \tag{1}$$

and the assignment of each region $R \in \Omega$ to a semantic label $l_R$. Unlike the semantic segmentation task, here a region should not contain more than one object instance of the same class. A region, however, may not be contiguous since occlusions can break regions into disconnected segments.

Recently, the supervised *detect-and-segment* approach has become increasingly popular due to its simplicity. First, a deep-learning based object detector is applied to the input image to generate $m$ detections in terms of bounding boxes $\mathcal{D}$. Then, a semantic segmentation network is applied to each bounding box to generate a segmentation mask for each instance, resulting in $m$ regions $\{R_1, R_2, \dots, R_m\}$. However, it is clear that the condition in (1) is not necessarily satisfied because

$$\cup_{i=1}^{m} R_i \subseteq \Omega, \quad R_i \cap R_j = \emptyset, \quad \neg \forall i \neq j. \tag{2}$$

This means that not all pixels in the image are segmented and two segmentation masks can overlap. While the second problem can be resolved using a pixel voting mechanism, the first problem is more challenging to be addressed. In *open-set* world, an image might capture objects that are unknown to the detector, so pixels belonging to these unknown object instances will not be labelled by this *detect-and-segment* approach. Miss-detected objects are not segmented either.

Ideally, one needs a model that is able to segment all individual objects (and "stuff") in an image regardless of whether they have been detected or not. In other words, all known and unknown object instances should be segmented. However, unknown and miss-detected objects will be assigned an "unknown" label.

Toward that goal, in this work, we propose a segmentation model that performs image segmentation globally (i.e., guaranteeing the condition $\cup_{i=1}^{k} R_i = \Omega$) so that each $R_i$ is a coherent region. The segmentation process also optimally assigns labels to these regions using the detection set $\mathcal{D}$. In the next section, we discuss our Bayesian formulation to achieve this goal.

## 4 Bayesian Formulation

Similar to the unsupervised Bayesian image segmentation formulation in [27], our image segmentation solution $S$ has the following structure:

$$S = ((R_1, t_1, \theta_1), (R_2, t_2, \theta_2), \ldots, (R_k, t_k, \theta_k)), \qquad (3)$$

where each region $R_i$ is "explained" by a model type $t_i$ with parameters $\theta_i$. More precise definitions of $t_i$ and $\theta_i$ will be given below. The number of regions $k$ is also unknown. In a Bayesian framework, the quality of a segmentation $S$ is measured as the density of a posterior distribution:

$$p(S|\mathbf{I}) \propto p(\mathbf{I}|S)p(S) \quad S \in \mathcal{S}, \qquad (4)$$

where $p(\mathbf{I}|S)$ is the likelihood and $p(S)$ is the prior, and $\mathcal{S}$ is the solution space. In the following, we will discuss the likelihood and prior terms used in our work.

### 4.1 The Likelihood Models

We assume that object regions in the image are mutually independent, forming the following likelihood term:

$$p(\mathbf{I}|S) = \prod_{i=1}^{k} p(\mathbf{I}_{R_i}|t_i, \theta_i). \qquad (5)$$

The challenge is to define a set of robust image models that explain complex visual patterns of object classes. The standard machine learning approach is to learn an image model for each object category using training images that have been manually annotated (i.e., segmented). Unfortunately, in open-set problems, as the number of object categories increases boundlessly, manually annotating training data for all possible object classes becomes infeasible.

In this work, we consider three types of image models to explain image regions: boundary/contour model ($\mathcal{C}$), bounding box model ($\mathcal{B}$), and mask model ($\mathcal{M}$) i.e., $t \in \{\mathcal{C}, \mathcal{B}, \mathcal{M}\}$. We use the boundary to describe unknown regions. More complicated models such as Gaussian mixture could also be used, but they have higher computational cost. The bounding box and mask models are used for known objects.

**Boundary/contour model ($\mathcal{C}$).** Objects in the image are often isolated by their contours. Assume that we have a method (e.g., COB [12]) that is able to estimate a contour probability map from the image. Given a region $R$, we can define its external boundary score $c_{ex}(R)$ as the lowest probability on the boundary, whereas its internal boundary score $c_{in}(R)$ is highest probability among internal pixels. The likelihood of the region $R$ being an object is defined as:

$$p(\mathbf{I}_R | c_{ex}(R), c_{in}(R)) \propto \left[\exp\left(-\frac{|c_{ex}(R) - 1|^2}{\sigma_{ex}^2}\right) \times \exp\left(-\frac{|c_{in}(R) - 0|^2}{\sigma_{in}^2}\right)\right]^{|R|} \quad (6)$$

where $\sigma_{ex}$ and $\sigma_{in}$ are standard deviation parameters. According to (6), a region with strong external boundary ($\approx 1$) and weak internal boundary ($\approx 0$) is more likely to represent an object. We used $\sigma_{in} = 0.4$ and $\sigma_{ex} = 0.6$.

**Bounding box model ($\mathcal{B}$).** Given an object detection $\mathbf{d}$ represented by a bounding box $\mathbf{b} = [c_x, c_y, w, h]$, object class $c$, and detection score $s$, the likelihood of a region $R$ being from the object $\mathbf{d}$ is:

$$p(\mathbf{I}_R | \mathbf{b}) \propto \text{IoU}(\mathbf{b}_R, \mathbf{b}) \times s \times \prod_{v \in R} \exp\left(-\frac{|v_x - c_x|^2}{\sigma_w^2}\right) \exp\left(-\frac{|v_y - c_y|^2}{\sigma_h^2}\right) \quad (7)$$

where $\mathbf{b}_R$ is the minimum bounding box covering the region $R$, $\text{IoU}(.)$ computes the intersection-over-union between two bounding boxes, $[v_x, v_y]$ is the location of pixel $v$ in the image space. $\sigma_w$ and $\sigma_h$, standard deviations from the center of the bounding box, are functions of bounding box width $w$ and height $h$ respectively. To avoid bigger bounding boxes with higher detection scores taking all the pixels, we encourage smaller bounding boxes by setting $\sigma_w = w^\alpha$ and $\sigma_h = h^\alpha$, where $\alpha$ is a constant smaller than 1. In our experiments, we set $\alpha = 0.8$.

**Mask model ($\mathcal{M}$).** Similarly, given an object detection $\mathbf{d}$ represented by a segmentation mask $\mathbf{m}$, object class $c$, and detection score $s$, the likelihood of a region $R$ being from the object $\mathbf{d}$ is:

$$p(\mathbf{I}_R | \mathbf{m}) \propto [\text{IoU}(R, \mathbf{m}) \times s]^{|R|}, \quad (8)$$

where $\text{IoU}()$ computes the intersection-over-union between two regions. Note that the mask model is optional in our framework.

### 4.2 The Prior Model

Our prior segmentation model is defined as:

$$p(S) \propto \exp(-\gamma k) \times \prod_{i=1}^{k} \exp\left(-|R_i|^{0.9}\right) \times \exp\left(-\rho(R_i)\right), \qquad (9)$$

where $k$ is the number of regions, and $\gamma$ is a constant parameter. In (9), the first term $\exp(-\gamma k)$ penalizes the number of regions $k$, and the second term $\exp(-|R_i|^{0.9})$ encourages large regions. The function $\rho(R_i)$, calculating the ratio of the total number of pixels in the region $R$ and the area of its convex hull, encourages compact regions. In our experiments, we set $\gamma = 100$.

## 5 MAP Inference using Simulated Annealing

Having defined the model for the semantic instance segmentation problem, the next challenge is to quickly find an optimal segmentation $S^*$ that maximizes the posterior probability over the solution space $\mathcal{S}$

$$S^* = \underset{S \in \mathcal{S}}{\operatorname{argmax}} \, p(S|\mathbf{I}), \qquad (10)$$

or analogously minimizing the energy $E(S, \mathbf{I}) = -\log(p(S|\mathbf{I}))$. The segmentation $S$ defined in (3) can be decomposed as $S = (k, \pi_k, (t_1, \theta_1), (t_2, \theta_2), \ldots, (t_k, \theta_k))$, where $\pi_k = (R_1, R_2, \ldots, R_k)$ is a partition of the image domain $\Omega$ into exactly $k$ non-overlap regions. Given a partition $\pi_k$, it is easy to compute the optimal $t_i$ and $\theta_i$ for each region $R_i \in \pi_k$ by comparing the likelihoods of $R_i$ given different image models. However, the more difficult part is the estimation of the partition $\pi_k$. Given an image domain $\Omega$, we can partition it into a minimum of 1 region and maximum of $|\Omega|$ regions. Let $\omega_{\pi_k}$ be the set of all possible partitions $\pi_k$ of the image into $k$ regions, then the full partition space is:

$$\mathcal{P} = \cup_{k=1}^{|\Omega|} \omega_{\pi_k}. \qquad (11)$$

It is clearly infeasible to examine all possible partitions $\pi_k$ with different values of $k$. We mitigate this problem by resorting to the Simulated Annealing (SA) optimization approach [28] to approximate the global optimum of the energy function $E(S, \mathbf{I})$.

### 5.1 Simulated Annealing

Algorithm 1 details our simulated annealing approach to minimizing the energy function $E(S, \mathbf{I}) = -\log(p(S|\mathbf{I}))$. Our algorithm performs a series of "moves" between image partitions ($\pi_k \to \pi_{k'}$) of different $k$ to explore the complex partition space $\mathcal{P}$, defined in (11). The model parameters $(t_i, \theta_i)$ for each region $R_i$ are computed deterministically at each step. A proposed segmentation is accepted probabilistically in order to avoid local minima.

8       T. Pham, V. Kumar B G, T-T Do, G. Carneiro, I. Reid**Algorithm 1** Simulated Annealing for Open-set Bayesian Instance Segmentation

**Input:** A set of detections (bounding boxes or masks), initial segmentation $S$, $E(S, \mathbf{I})$, and temperature $T$.
**Output:** Optimal segmentation $S^*$.
1: $S^* = S$.
2: Sample a neighbor partition $\pi_{k'}$ near the last partition $\pi_k$.
3: Update parameters $(t_i, \theta_i)$ $i = 1, 2, \ldots, k'$.
4: Create a new solution $S = (k', \pi_{k'}, (t_1, \theta_1), \ldots, (t_{k'}, \theta_{k'}))$.
5: Compute $E(S, \mathbf{I})$
6: With probability $\exp\left(\frac{E(S^*, \mathbf{I}) - E(S, \mathbf{I})}{T}\right)$, $S^* = S$.
7: $T = 0.99T$ and repeat from Step 2 until the stopping criteria is true.

A crucial component of Algorithm 1 is the sampling of new partition $\pi_{k'}$ near by the current partition $\pi_k$ (Line 2). The sooner good partitions are sampled, the faster Algorithm 1 reaches the optimal $S^*$. In Section 5.2, we propose an efficient partition sampling method based on a region hierarchy.

### 5.2 Efficient Partition Sampling

The key component of our Simulated Annealing based instance segmentation approach is an efficient image partition generator based on a boundary-driven region hierarchy.

**Boundary-driven region hierarchy** A region hierarchy is a multi-scale representation of an image, where regions are groups of pixels with similar characteristics (i.e., colors, textures). Similar regions at lower levels are iteratively merged into bigger regions at higher levels. A region hierarchy can be efficiently represented using a single Ultrametric Contour Map (UCM) [1]. A common way to construct an image region hierarchy is based on image boundaries, which can be either estimated using local features such as colors, or predicted using deep convolutional networks (e.g., [12]). In this work, we use the COB network proposed in [12] for the object boundary estimation due to its superior performance compared to other methods.

Let $\mathcal{R}$ denote the region hierarchy (tree). One important property of $\mathcal{R}$ is that one can generate valid image partitions by either selecting various levels of the tree or performing tree cuts [14]. Conditioned on $\mathcal{R}$, the optimal tree cut can be found exactly using Dynamic Programming, as done in [14]. Unfortunately, regions of the hierarchy $\mathcal{R}$ might not represent accurately all complete objects in the image due to imperfect boundary estimation. Also, occlusion might cause objects to split into different regions of the tree. As a result, the best partition obtained by the optimal tree cut may be far away from the optimal partition $\pi_k^*$. Below, we show how to sample higher-quality image partitions based on the initial region hierarchy $\mathcal{R}$.

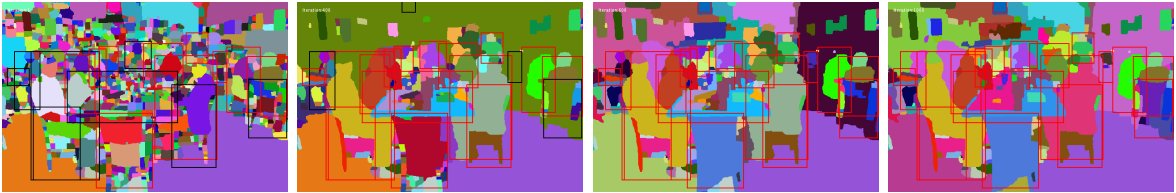

**Fig. 2:** Intermediate segmentation results when the Algorithm 1 progresses. Left is the initialized segmentation. Right is the final result when the algorithm converges. In each image, bounding boxes represent detected objects returned by the trained detector. Notice black bounding boxes are currently rejected by the algorithm

**Image partition proposal** Let $\pi_k = (R_1, R_2, \ldots, R_k) \subset \mathcal{R}$ be the current image partition, a new partition can be proposed by first randomly sampling a region $R \in \mathcal{R} \setminus \pi_k$, then "paste" it onto the current partition $\pi_k$. Let $\mathcal{A}_R \subset \pi_k$ be a subset of regions that overlap with $R$, where $|\mathcal{A}_R|$ denotes the number of regions in $\mathcal{A}_R$. The following scenarios can happen:

- $R = \cup \mathcal{A}_R$. Regions in $\mathcal{A}_R$ will be merged into a single region $R$.
- $|\mathcal{A}_R| = 1, R \subset \mathcal{A}_R$. $\mathcal{A}_R$ will be split into two subregions: $R$ and $\mathcal{A}_R \setminus R$.
- $|\mathcal{A}_R| > 1, R \subset \cup \mathcal{A}_R$. Each region in $\mathcal{A}_R$ will be split by $R$ into two subregions, one of which will be merged into $R$. This is a split-and-merge process.

It can be seen that the above "sample-and-paste" operation naturally realizes the split, merge, and split-and-merge processes probabilistically, allowing the exploration of partition spaces of difference cardinalities. Note that the last two moves may generate new region candidates that are not in the original region hierarchy $\mathcal{R}$. These regions are added into $\mathcal{R}$ in the next iteration. Figure 2 demonstrates the progressive improvement of the segmentation during Simulated Annealing optimisation.

**Occlusion handling** The above "sample-and-paste" process is unlikely to be able to merge regions that are spatially separated. Because of occlusion, spatially isolated regions might be from the same object instance. Given a current partition $\pi_k$ and a detection represented by either a bounding box **b** or a mask **m**, we create more region candidates by sampling pairs of regions in $\pi_k$ that overlap with **b** or **m**. These regions are added into $\mathcal{R}$ in the next iteration.

## 6 Experimental Evaluation

In all below experiments, we run the Algorithm 1 for 3000 iterations. For each image, we run the COB network [12] and compute a region hierarchy of 20 levels, in which level 10 will be used as the initialized segmentation.

### 6.1 Baselines

Since we are not aware of any previous work solving the same problem as ours, we develop a simple baseline for comparisons. Noting that the input to our method

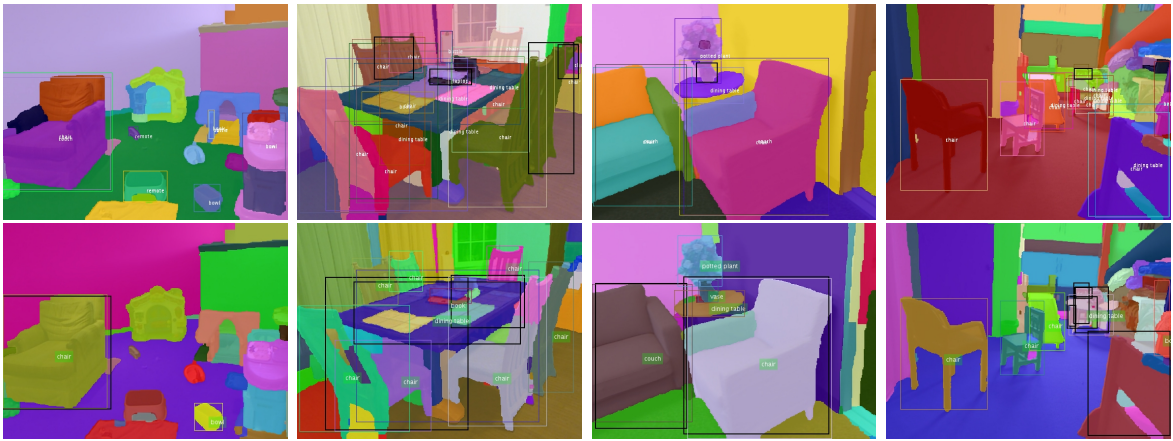

**Fig. 3:** Baseline (top row) vs our method (second row) with bounding box supervision. Testing images are from the NYU dataset. Bounding boxes represent detected objects. Note that not all detected object instances are used in the final segmentation. Black bounding boxes are detections rejected by the methods

is an image, and possibly a set of object detections or masks returned either by an object detection (e.g., Faster-RCNN) or an instance segmentation method (e.g., Mask-RCNN) trained on known classes. In some cases, no known objects are detected in the image. For the baseline method, we first apply an unsupervised segmentation method to decompose the image into a set of non-overlap regions. If a set of detections (bounding boxes) is given, we classify each segmented region into these detected objects using intersection-over-union scores. If the maximum score is smaller than 0.25, we assign that region to an unknown class. When a set of object masks is given, we overwrite these masks onto the segmentation. Masks are sorted (in ascending order) using detection scores to ensure that high confidence masks will be on top.

We develop the unsupervised segmentation by thresholding the UCMs, computed from the boundary maps estimated by the COB network [12]. We use different thresholds for the baseline method, including the best threshold computed using ground-truth data. As reported in [12,14], this segmentation method greatly outperforms other existing unsupervised image segmentation methods, making it a strong baseline for comparison.

### 6.2  Open-set Datasets

For evaluation, we create a testing environment which includes both known and unknown object classes. In computer vision, the COCO dataset has been widely used for training and testing the object detection and instance segmentation methods. This dataset has annotations (bounding boxes and masks) of 80 object classes. We select these 80 classes as known classes. Moreover, the popular NYU dataset has annotations of 894 classes, in which 781 are objects and 113 are stuffs. We observe (manually check) that 60 classes from the COCO dataset actually appear in the NYU dataset. Consequentially, we select the NYU dataset as the

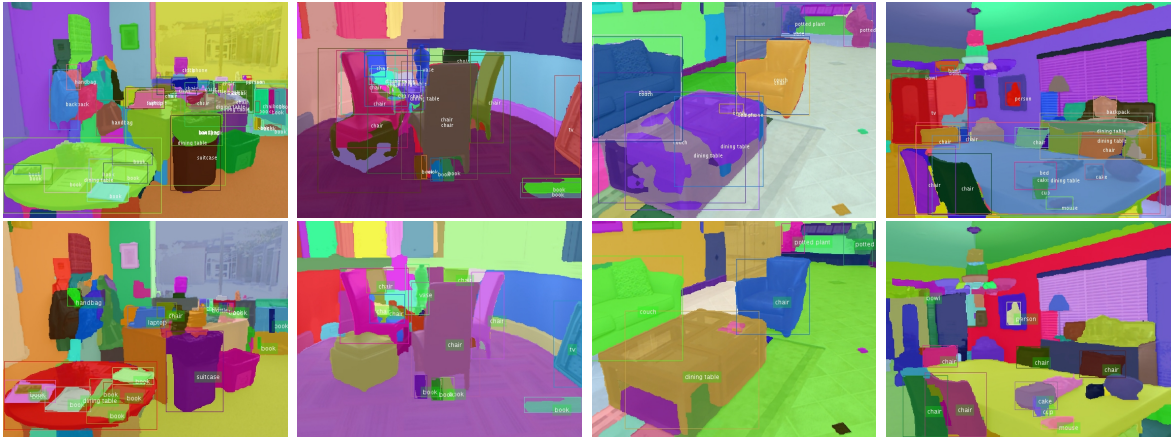

**Fig. 4:** Baseline (top row) vs our method (second row) with mask supervision. Testing images are from the NYU dataset. Bounding boxes represents detected objects

testing set with 60 known and 721 unknown for benchmarking our method and baseline method.

### 6.3 Ablation Studies

We compare our method against the baseline in three different settings: 1) No supervision, 2) Bounding box supervision and 3) Mask supervision. In the first case, we assume that there is no training data available for training the object detection or instance segmentation networks. In the second case, we assume that known object classes are annotated with only bounding boxes so that one can train an object detector (i.e., Faster-RCNN). It is worth mentioning that while our method can be guided by a given set of bounding boxes (if available), the baseline method does not use the given bounding boxes for segmentation at all because the object segmentation and object labeling are carried sequentially. Finally, in the last setting, if known object instances are carefully annotated with binary masks, one can train an instance segmentation network (i.e., Mask-RCNN), which is then applied onto testing images to return a set of segmentation masks together with their categories. The predicted segmentation masks are taken as input to the baseline and our method. In all our experiments, we use Detectron[1], which implements Mask-RCNN method, to generate bounding boxes and segmentation masks. We select the model trained on the COCO dataset.

**Evaluation** For each image, we first run the Hungarian matching algorithm to associate ground truth regions to predicted regions based on IoU scores. We then compute, given an IoU threshold, precision and recall rates, which will be summarised via F-1 scores. Note that we evaluate known and unknown object classes separately. Table 1 reports comparison results tested on NYU images using F-1 scores at different IoU thresholds. Firstly, it is clear that our method performs much better than the baseline when both methods are not guided by detections, even when the baseline is provided the best threshold (0.4) computed using ground truth. Moreover, when guided by bounding boxes and masks, our accuracies on known object classes increase significantly as expected. In contrast, the baseline method's accuracies decrease greatly when masks are used because the given masks are greedily overwritten onto the unsupervised segmentation results. These results confirm the efficacies of our global Bayesian image segmentation approach compared to the greedy baseline method.

---

[1] https://github.com/facebookresearch/Detectron

Table 1: Quantitative comparison results on 654 NYU RGB-D testing images between our method and the baseline method with different supervision information. The baseline method is tested with different thresholds. We report F-1 scores for known and unknown classes at 0.5 and 0.75 IoU thresholds respectively

| Method | Supervision | Known | | Unknown | |
|---|---|---|---|---|---|
| | | $F_1^{50}$ | $F_1^{75}$ | $F_1^{50}$ | $F_1^{75}$ |
| Baseline (0.3) | None/BBoxes | 40.1 | 21.1 | 47.8 | 26.3 |
| Baseline (0.3) | Masks | 10.6 | 5.1 | 19.5 | 10.9 |
| Baseline (0.4) | None/BBoxes | 47.4 | 26.1 | 45.2 | 26.7 |
| Baseline (0.4) | Masks | 7.3 | 3.8 | 13.25 | 7.9 |
| Our method | None | 45.6 | 22.6 | 55.7 | 32.2 |
| Our method | BBoxes | 48.6 | 23.1 | 54.2 | 30.4 |
| Our method | Masks | 51.1 | 25.9 | 53.8 | 30.3 |

Table 2: Comparison results on 80 known classes tested on 5k COCO validation images. $mIoU_w$ is weighted by the object sizes

| Method | Supervision | mAP | $mIoU_w$ | mIoU |
|---|---|---|---|---|
| Baseline | Weakly (Boxes) | 10.1 | 26.6 | 25.2 |
| Our method | Weakly (Boxes) | 20.0 | 33.6 | 32.3 |
| Mask-RCNN | Fully (Boxes and Masks) | 30.5 | 38.7 | 37.3 |

Figures 3 and 4 demonstrate the qualitative comparison results between our method and the baseline. It can be seen that the baseline method fails to segment objects correctly (either under-segmentation or over-segmentation). In contrast, our method, guided by the given bounding boxes, performs much better. More importantly, the baseline method does not take the given bounding boxes into segmentation, it can not suppress multiple duplicated detections (with different classes) at the same location, unlike our method.

### 6.4   Weakly Supervision Segmentation of Known Objects

Existing instance segmentation methods (e.g., Mask-RCNN) require ground-truth instance masks for training. However, annotating segmentation masks for all object instances is very expensive. Nonetheless, our semantic instance segmentation method does not require mask annotations for training. Here, we compare our weakly supervision instance segmentation of known objects against the fully supervised Mask RCNN method. Recently, Hu et al. [8] have proposed a learning transfer method, named Mask$^X$RCNN, for instance segmentation when only a subset of known object classes has mask annotations. We are, however, unable to compare with Mask$^X$RCNN as neither its pre-trained model nor predicted segmentation masks are publicly available.

**Evaluation** While our method outputs one instance label per pixel, Mask RCNN returns a set of overlap segmentation masks per image. Therefore, the two methods can be not practically compared. To be fair, we post-process the Mask RCNN's results to ensure that one pixel is assigned to only one instance (via pixel voting based on detection scores). We measure the segmentation accuracies using Mean Intersection over Union (mIoU) metric. We first run the Hungarian matching algorithm to match predicted regions to ground-truth regions. The "matched" IoU scores are then averaged over all object instances and semantic categories. We also report Mean Average Precision (mAP) scores as Mask RCNN does. However, we note that mAP metric is only suitable for problems where an output is a set of ranked items. In contrast, our method returns, for each image, a single pixelwise segmentation where each pixel is assigned to a single object instance without any ranking.

Table 2 reports the comparison results. It can be seen that our method, though only requiring bounding box supervision, is competitive with respect to Mask RCNN, which requires ground-truth segmentation masks of all known object instances for training. This again indicates the efficacy of our method for the *open-set* instance segmentation problem where it is expensive, if not impossible, to annotate segmentation masks for all object instances. Figure 5 demonstrates example semantic instance segmentation results from our method using images from COCO dataset. Notice that our method is not only able to segment known objects but also unknown objects and stuffs such as grass, sky with high accuracies.

## 7  Discussion and Conclusion

We have presented a global instance segmentation approach that has a capability to segment all object instances and stuffs in the scene regardless of whether these objects are known or unknown. Such a capability is useful for autonomous robots working in *open-set* conditions [23], where the robots will unavoidably encounter novel objects that were not part of the training dataset.

Different from state-of-the-art supervised instance segmentation methods [4,7,19,29], our approach does not perform segmentation on each detection independently, but instead segments the input image globally. The outcome is a set of coherent regions which are perceptually grouped and are each associated either to a known detection or unknown object instance. We formulate the instance

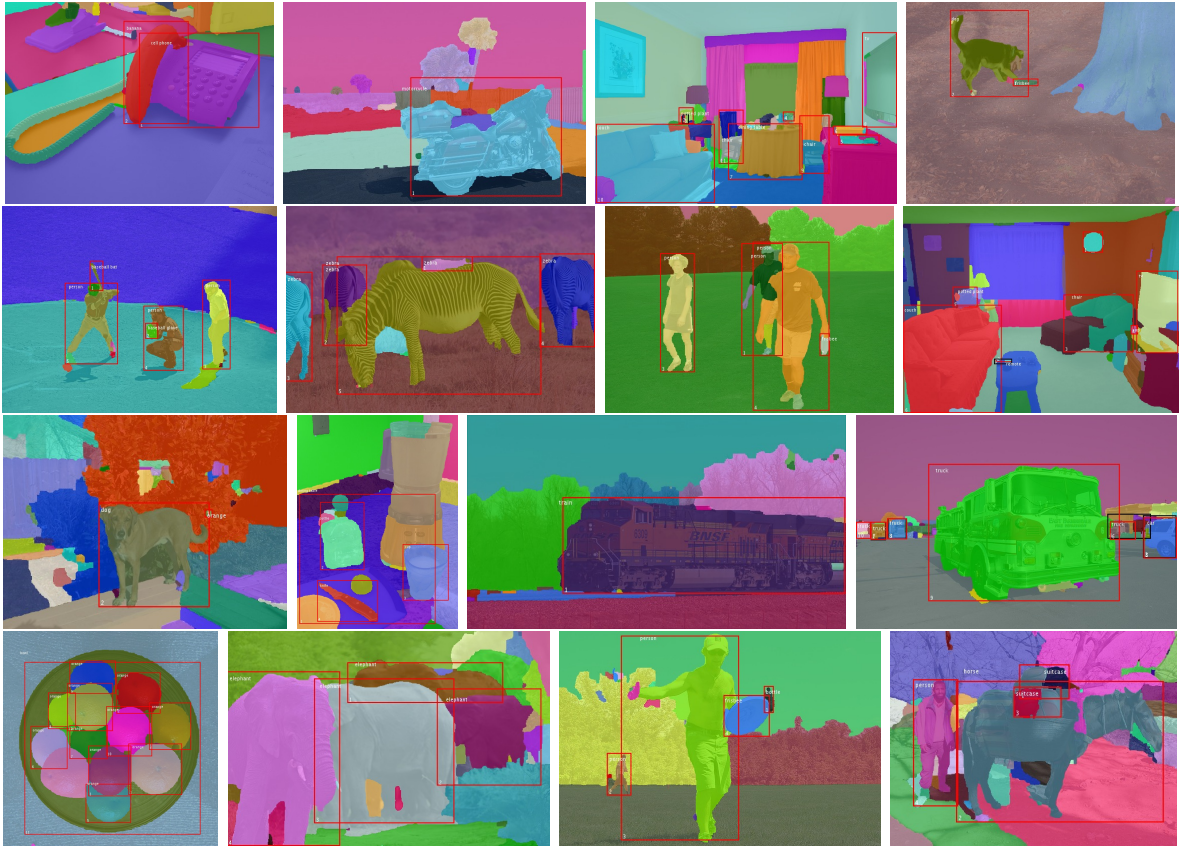

**Fig. 5:** Example instance segmentation results of our method on COCO dataset. Bounding boxes represents detected objects. In these examples, our method only uses bounding box supervision. Notice that our method segments not only detected objects, but also other miss-detected and unknown objects

segmentation problem in a Bayesian framework, and approximate the optimal segmentation using using a Simulated Annealing approach.

We envision that *open-set* instance segmentation will soon become a hot research topic in the field. We thus believe the proposed method and the experimental setup proposed will serve as a strong baseline for future methods to be proposed in the field (e.g., end-to-end learning mechanisms).

Moreover, existing supervised learning methods which require a huge amount of precise mask annotations for all object instances for training, which is very expensive to extend to new object categories. Our approach offers an alternative, which is based on a more natural incremental annotation strategy to deal with new classes. This strategy consists of explicitly identifying unknown objects from images and training new object models using the labels provided by an "oracle" (such as a human).

**Acknowledgements** This research was supported by the Australian Research Council through the Centre of Excellence for Robotic Vision (CE140100016) and by Discover Project (DP180103232).